%% file: root.tex
\documentclass[letterpaper, 10 pt, journal, twoside]{IEEEtran}

\IEEEoverridecommandlockouts                              

\makeatletter
\let\NAT@parse\undefined
\makeatother
\usepackage[numbers]{natbib}

\usepackage[pdftex]{graphicx}
\usepackage{amsmath}
\usepackage{amssymb}  
\usepackage{multirow}
\usepackage{array,booktabs}
\usepackage{diagbox}
\usepackage{balance}
\usepackage[ruled]{algorithm}
\usepackage{algpseudocode}
\usepackage{caption}
\usepackage{subcaption}
 \usepackage[final]{hyperref}
 \hypersetup{
     colorlinks=true,
     linkcolor=blue,
     filecolor=magenta,
     urlcolor=blue,
     citecolor=black
 }
\usepackage{./rpm_packages/rpm_SIunits}
\usepackage{./rpm_packages/rpm_acronyms}
\usepackage{./rpm_packages/rpm_math}
\usepackage{./rpm_packages/rpm_misc}

\usepackage{soul,color}
\usepackage{lipsum}

\usepackage{xcolor}

\newcommand{\bl}[1]{{\textcolor{black}{#1}}}

\usepackage{tikz} 

\title{Thermal Chameleon: Task-Adaptive Tone-mapping for Radiometric Thermal-Infrared images}

\author{Dong-Guw Lee${}^{1}$, Jeongyun Kim${}^{1}$, Younggun Cho${}^{2}$, and Ayoung Kim${}^{1*}$

\thanks{Manuscript received: July, 31, 2024; October, 7, 2024.}
\thanks{This paper was recommended for publication by
Editor Markus Vincze upon evaluation of the Associate Editor and Reviewers’ comments.
This work was jointly supported by the New Faculty Startup Fund from Seoul National University, the Korea Agency for Infrastructure Technology Advancement (KAIA) grant funded by the Ministry of Land, Infrastructure and Transport (Grant RS-2023-00250727) through the Korea Floating Infrastructure Research Center at Seoul National University. and the National Research Foundation of Korea(NRF) grant funded by the Korea government(MSIT) (No. RS-2023-00302589)}

\thanks{$^{1}$D. Lee, J. Kim, and A. Kim are with the Department of Mechanical Engineering, SNU, Seoul, S. Korea {\tt\small [donkeymouse, jeongyunkim, ayoungk]@snu.ac.kr}}%
\thanks{$^{2}$Y. Cho is with the Department of Electrical Engineering, Inha University, Incheon, S. Korea {\tt\footnotesize yg.cho@inha.ac.kr}}%
\thanks{Digital Object Identifier (DOI): see top of this page.}
}

\begin{document}

\maketitle

\input{abstract}
\input{introduction}

\input{relatedwork}

\input{method}
\input{experiment}

\input{conclusion}
\balance
\small
\bibliographystyle{IEEEtranN} 
\bibliography{string-short,references,output}

\end{document}

%% file: abstract.tex
\begin{abstract}

Thermal Infrared (TIR) imaging provides robust perception for navigating in challenging outdoor environments but faces issues with poor texture and low image contrast due to its 14/16-bit format. Conventional methods utilize various tone-mapping methods to enhance contrast and photometric consistency of TIR images, however, the choice of tone-mapping is largely dependent on knowing the task and temperature dependent priors to work well. In this paper, we present Thermal Chameleon Network (TCNet), a task-adaptive tone-mapping approach for RAW 14-bit TIR images. 
Given the same image, TCNet tone-maps different representations of TIR images tailored for each specific task, eliminating the heuristic image rescaling preprocessing and reliance on the extensive prior knowledge of the scene temperature or task-specific characteristics. TCNet exhibits improved generalization performance across object detection and monocular depth estimation, with minimal computational overhead and modular integration to existing architectures for various tasks. Project Page: https://github.com/donkeymouse/ThermalChameleon

\end{abstract}

%% file: introduction.tex
\section{Introduction}
\label{sec:intro}

\ac{TIR} images operate in the long wave infrared spectrum, perceiving the environment through temperature. This characteristic ensures robust perception in adverse environments with poor illumination and airborne obscurants, making \ac{TIR} sensors desirable for robotics \cite{shin2019sparse} and computer vision \cite{shin2023wacv}. However, since \ac{TIR} images are usually stored in 14 or 16-bit format, such high dynamic range results in narrow image distribution, yielding ambiguous object boundaries and poor image contrasts \cite{hadar}.

\begin{figure}[t]
  \centering
  \begin{subfigure}{\columnwidth}
    \centering
    \includegraphics[width=\columnwidth]{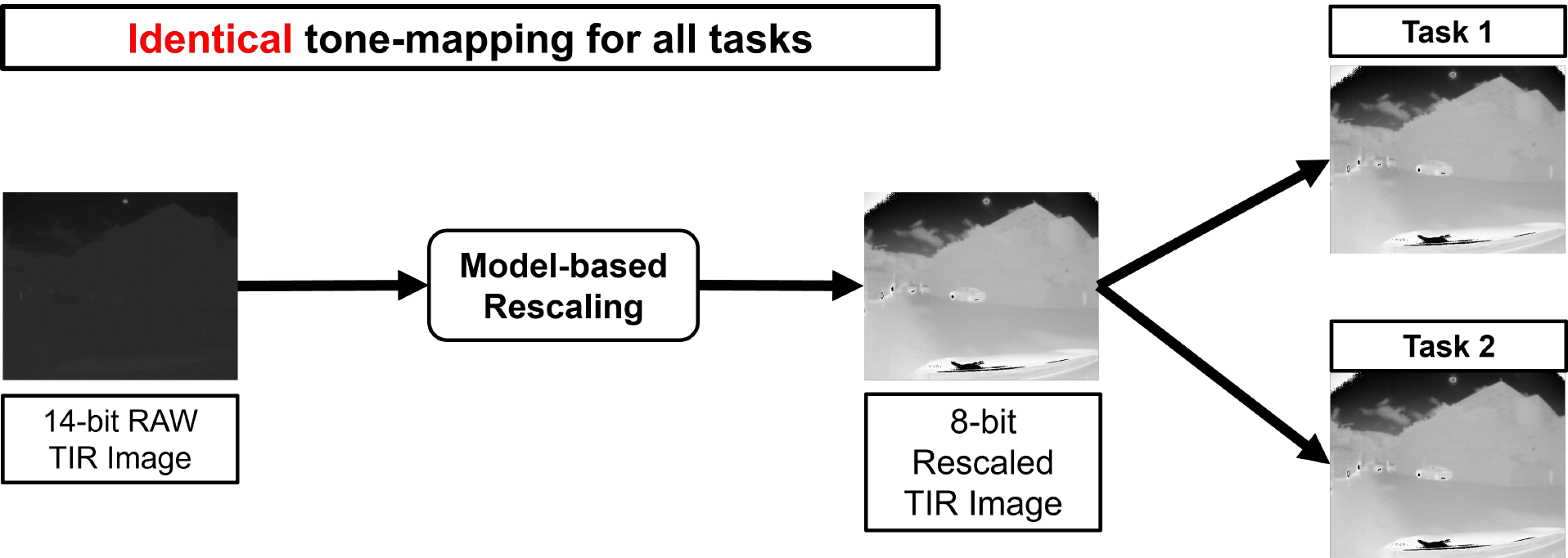}
    \caption{Conventional TIR tone-mapping: Identical tone-mapping used for all tasks.}
    \label{conventional_methods}
  \end{subfigure}
  \hfill
  \begin{subfigure}{\columnwidth}
    \centering
    \includegraphics[width=\columnwidth]{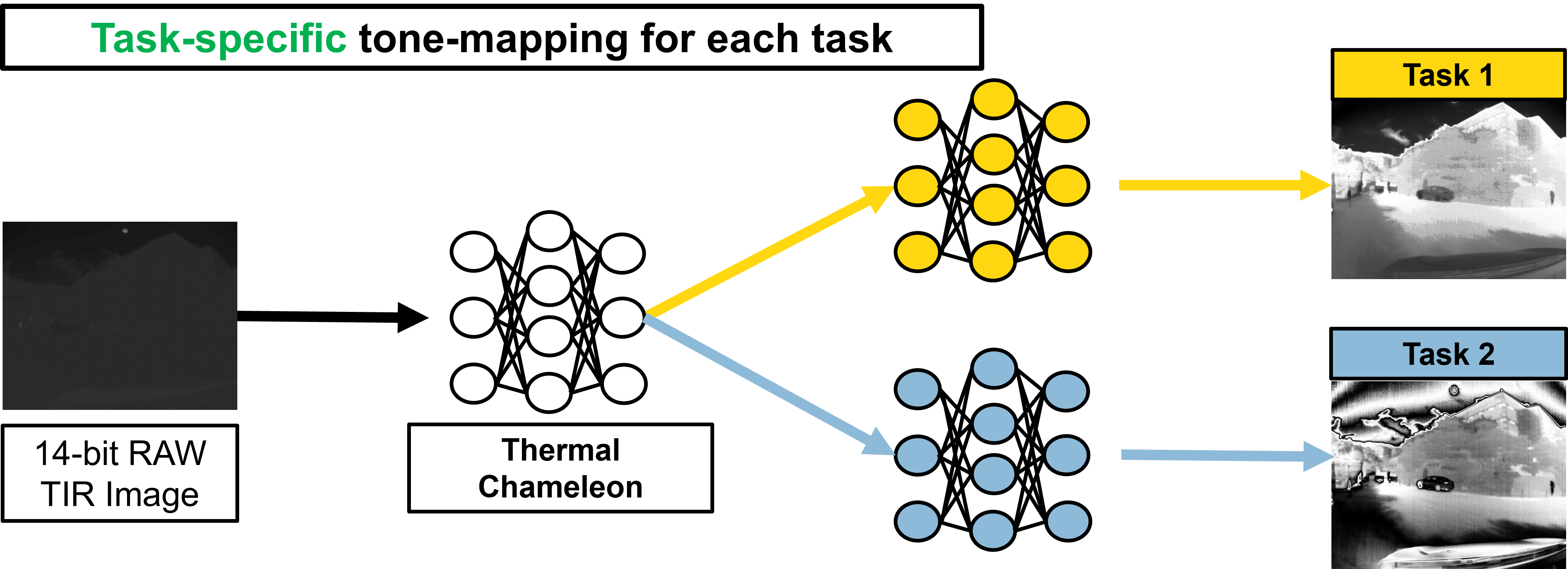}
    \caption{Thermal Chameleon Net is a learning-based network that task-adaptively tonemaps \ac{TIR} images optimized for each task. }
    \label{tcnet_method}
  \end{subfigure}
  \caption{(a): Conventional methods use a single TIR representation for all tasks (b): TCNet is a learning-based tone-mapping network for \ac{TIR} images. Given the same network, TCNet achieves different tone-mapping, tailored for each task.}

  \label{main_fig_overview}
\end{figure}

A common way to address such issue is by linearly rescaling \ac{TIR} images with their minimum and maximum pixel values, however, this method performs poorly on long tailed image distribution caused by hot or cold objects \cite{khattak2020keyframe}. Recent works have proposed improved tone-mapping methods to maintain photometric consistency between consecutive frames \cite{shin2023wacv, das2021online} or use non-linear gains to enhance contrast \cite{gil2024fieldscale, shin2022raliros, godrich2023spie}. Nonetheless, as shown in \figref{conventional_methods}, existing methods apply identical tone-mapping for all \ac{TIR} images. Thus, users must rely on extensive domain knowledge and heuristics to select the appropriate tone-mapping for a given task, requiring excessive handcrafting and familiarity with the temperature data.


In this work, we propose \ac{TCNet}, a task-adaptive tone-mapping network for radiometric \ac{TIR} images. Inspired by \cite{xu2023toward}, which replaced conventional image processing with neural network-based tone-mapping for raw RGB images, TCNet similarly replaces model-based rescaling with neural network-based tone-mapping (\figref{tcnet_method}). This approach achieves task-optimized tone-mapping, eliminating extensive preprocessing. Unlike recent learning-based \ac{TIR} tasks that rely on preprocesseing a single identical \ac{TIR} image, TCNet integrates tone mapping within the network to produce different representations of the same \ac{TIR} image optimized for each task. This eliminates the need for users to know task-specific characteristics or scene temperature priors for choosing the adequate tone-mapping method.  

Our contributions are summarized as follows:

\begin{itemize}
    \item \textbf{Temperature-aware data representation and augmentation}: We propose thermal embedding, a data augmentation technique for radiometric \ac{TIR} images. Thermal embedding uses absolute per-pixel temperature as a common ground to create multiple \ac{TIR} representations from a single image. This simulates varying environmental temperatures without requiring extensive knowledge of the environment, improving the ability to represent high dynamic range \ac{TIR} images. This enhances data diversity, particularly for monocular image-based tasks.
   
   \item \textbf{Adaptive Channel Compression}: TCNet employs an adaptive channel compression to assign task-specific weights to each thermal embedding. This facilitates task-adaptive tone-mapping, promotes modular integration into various tasks, and reduces computational overhead and artifacts.

   \item \textbf{Task-adaptive tone-mapping}: TCNet formulates task-optimized tone-mapping for different tasks, eliminating the need for excessive preprocessing and domain knowledge to select optimal tone-mapping for unseen environment. Additionally, task-adaptive tone mapping provides insights into the image characteristics favored by each task.


\end{itemize}

%% file: relatedwork.tex
\section{related work}
\label{sec:relatedwork}
\subsection{\bl{TIR image tone-mapping} }


\bl{Tone-mapping in \ac{TIR} images, also known as image rescaling, usually refer to the process of converting raw 14-bit \ac{TIR} images into 8-bit format. Such process can be categorized into linear and non-linear rescaling.} For linear rescaling, previous works estimate adequate maximum and minimum rescaling parameters through TIR camera models \cite{das2021online}, temporal windows \cite{shin2021ralcolor}, or clipping bounds established at the 1st and 99th percentile of the image distribution \cite{shin2023wacv}. 


Non-linear rescaling enhances local and global contrasts of \ac{TIR} images. \citeauthor{shin2022raliros} \cite{shin2022raliros} employed variable bin-based CLAHE to standardize long-tailed distributions in \ac{TIR} images. JointTMO \cite{godrich2023spie} used multiscale Retinex transform with a tone-mapping network, but the performance of the learning-based tone-mapping is upper-bounded by the quality of the ground truths tone-mapped image used for training. Fieldscale \cite{gil2024fieldscale} proposed 2D grid-based min and max rescaling, however, temporal coherence between frame is not preserved. Overall, non-linear rescaling methods tend to attenuate local details within the less populated regions of the image and amplifies the noise inherent to \ac{TIR} images.

While most tasks focus on improving image preprocessing aspects of \ac{TIR} images, thereby yielding a single image representation for all tasks, we introduce the first task-dependent, learning-based tone-mapping network for \ac{TIR} images, allowing for self-optimized tone-mapping tailored to specific tasks.



\subsection{Advances in TIR Image Augmentation}
Traditional RGB-based tone-mapping applies variants of gamma correction to Bayer-filtered RGB images. However, these methods are designed for multiple luminance channels and are not suitable for \ac{TIR} images, which operate on a single luminance channel \cite{godrich2023spie}. Previous works have used image inversion \cite{herrmann2018cnn} to diversify \ac{TIR} images, but since then, only few advancements to \ac{TIR} image-based augmentation have emerged.

Learning-based methods use neural networks to estimate gamma correction parameters or refine images \cite{xu2023toward}. However, these methods often rely on Bayer filtering to standardize a single high dynamic range image into multiple representations, which is not feasible for \ac{TIR} images. Thermal embedding addresses this gap by representing a single raw high dynamic range \ac{TIR} image into multiple representations, facilitating learning-based tone-mapping for \ac{TIR} images.

\subsection{TIR image-based robotics applications} 

RAW TIR images have been prominently used in direct SLAM methods \cite{shin2019sparse, khattak2020keyframe} for their ability to retain photometric consistency. Ensuring this consistency between pairs of \ac{TIR} images is crucial for low-level vision tasks such as monocular depth estimation \cite{shin2021ralcolor, shin2022raliros, shin2023wacv} and pose estimation \cite{saputra2020deeptio}. However, due to low contrast in the \ac{TIR} images, many methods rely on supervision from other sensors such as LiDAR \cite{shin2019sparse} or RGB images \cite{saputra2020deeptio}. For object detection, non-linear rescaling is often used to maximize global contrast, though it can lead to image saturation \cite{gil2024fieldscale, pbvs_raw}.

Choosing the right rescaling method depends on the environmental context of the scene. Histogram equalization is commonly used for \ac{TIR} images with long-tailed distributions \cite{shin2022raliros}, while linear scaling is preferred for tasks requiring temporal consistency. Ultimately, optimal results demand extensive domain knowledge of the scene and task from the users.

\ac{TCNet} optimizes \ac{TIR} image processing independently, without the need for auxiliary supervision from other modalities. It uses task-dependent losses to create task-optimized \ac{TIR} images, eliminating the ambiguity of choosing the right tone mapping. Additionally, the adaptive channel compression module allows for integration into various existing architectures, supporting multiple downstream tasks.

%% file: method.tex
\section{Method}
\label{sec:method}

\begin{figure*}
  \centering 
\includegraphics[width=\textwidth]{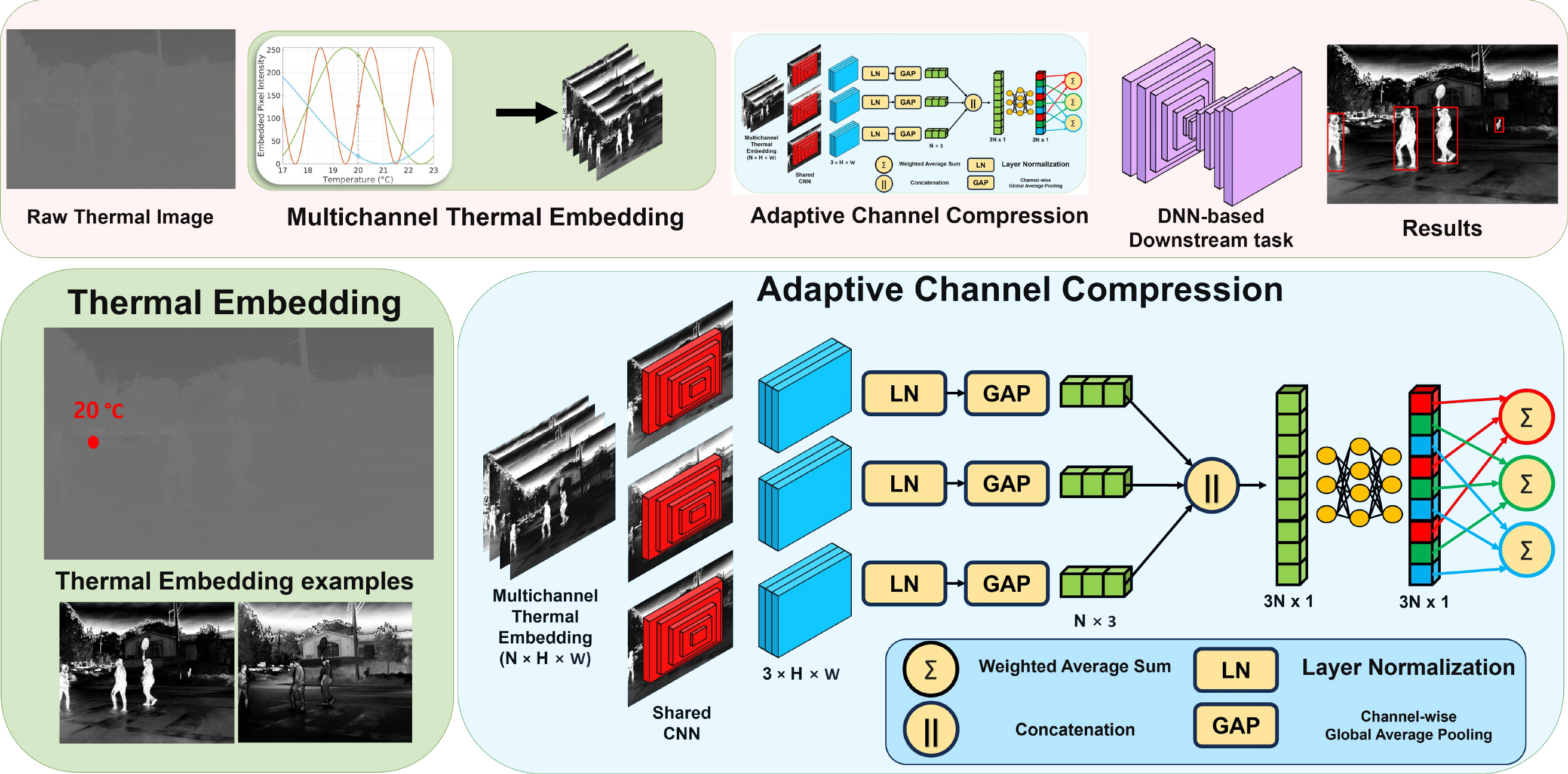}
\caption{Overview of Thermal Chameleon Network. The network consists of two stages: multichannel thermal embedding (green) which formulates diverse representations from a 14-bit radiometric \ac{TIR} images and an adaptive channel compression network (blue) that adaptively compresses these multichannel thermal embeddings into 3 channels which is then used to be trained on downstream tasks. \bl{All components are trained in an end-to-end manner.}}
\label{fig:network_architecture}

\end{figure*}

\begin{figure*}
  \centering 
\includegraphics[width=\textwidth]{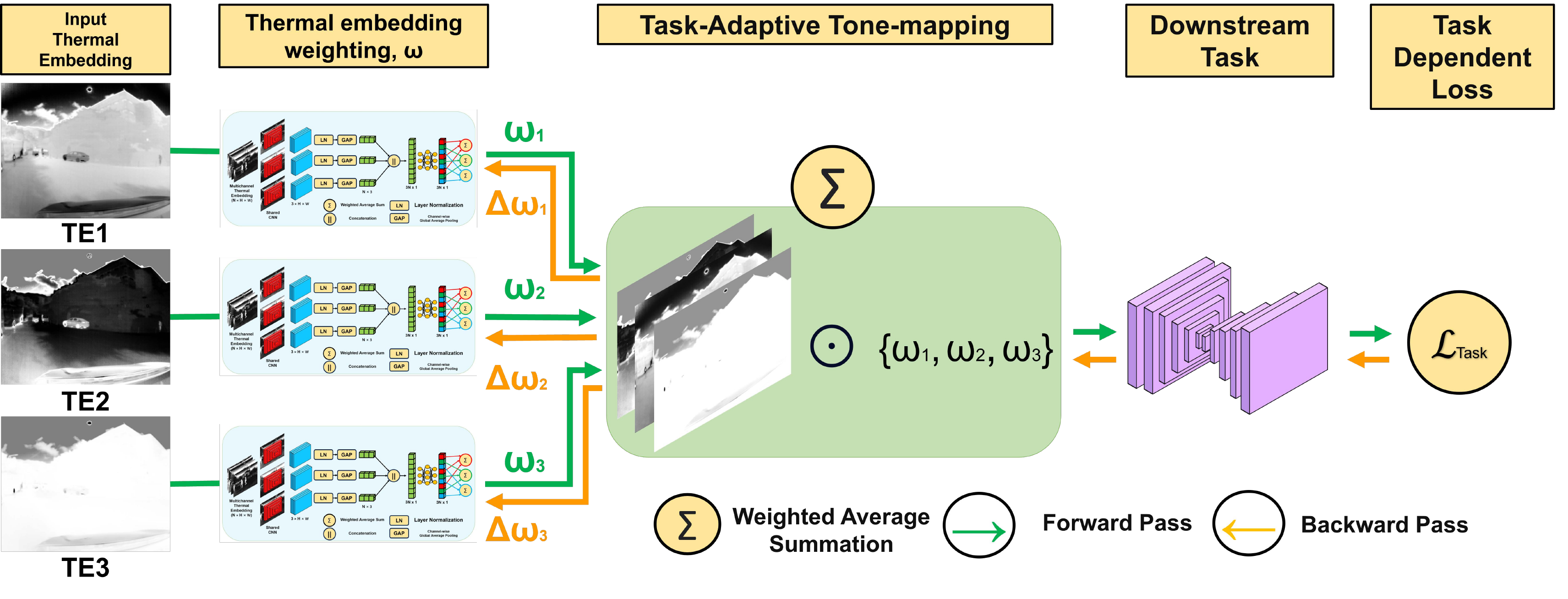}
\caption{Adaptive channel compression. Given multichannel \ac{TIR} embedding, the network assigns adaptive weightings for each thermal embedding optimized by the task-dependent loss. Using such weightings, we adaptively tone-map \ac{TIR} images via weighted average summation of each thermal embedding and the respective weights.}
\label{fig:adaptive_overview}

\end{figure*}

  


\subsection{Overview}




An overview of \ac{TCNet} is depicted in \figref{fig:network_architecture}. \ac{TCNet} comprises two stages: multichannel thermal embedding and adaptive channel compression. First, multichannel thermal embedding encodes a single-channel RAW \ac{TIR} image into multiple thermal embeddings, concatenated channel-wise (green). Then, adaptive channel compression extracts weights from each thermal embedding to compute task-adaptive tone-mapping, forming a three-channel tone-mapped feature (blue). Finally, this feature is propagated into the downstream task. 

\subsection{Multichannel Thermal Embedding}

  


  




\subsubsection{Absolute temperature conversion}
Different \ac{TIR} cameras have varying offsets and lens calibration parameters, leading to discrepancies in encoded pixel values. These differences cause distribution shifts due to temperature variations and camera model discrepancies. To address this, we convert 14-bit RAW \ac{TIR} images into absolute temperatures in degrees Celsius using formula driven from Planck's Law \cite{tlinear}. This standardization ensures a consistent representation of temperature across different cameras.
\begin{equation}
   I_{C} = \frac{P}{\ln\left(\frac{R}{S-O} + F\right)} - 273.15
\label{temp_conversion}
\end{equation}
where $P$ and $R$ are constants derived from speed of light, the Boltzmann constant, and Planck's constant, $S$ is the value of the 14-bit RAW \ac{TIR} image pixel, $O$ and $F$ are the \ac{TIR} camera's offset and calibration parameters. 

\subsubsection{Thermal embedding}

Thermal embedding projects a temperature image $I_{C}$ using a sinusoidal embedding $I_{ME}: \mathbb{R}^{B \times H \times W \times 1} \times \mathbb{N}^{1 \times N} \mapsto \mathbb{R}^{B \times H \times W \times N}$, as denoted by Equation \eqref{single_channel_function}

\begin{equation}
I_{ME} = \underset{i=1} {\overset{N} {\LARGE ||}} (\frac{255}{2} \cdot \sin{(\frac{I_{C}}{D_{i}}\cdot \pi}) + \frac{255}{2})
\label{single_channel_function}
\end{equation}


where $D$, $B$, $N$, $H$, $W$ and $\LARGE ||$ symbolize a set of temperature periodic constants, batch size, number of channels, image height, image width, and channel-wise concatenation of thermal embeddings respectively. Here, $D = \{D_1, D_2, \ldots, D_N\}$ in which each $D$ is used to map a single temperature to different intensities. We apply thermal embedding (\ref{single_channel_function}) with randomly selected $D_{i}$ to form a $N$-channel thermal embedding. 

The idea of sinusoidal function for embedding temperature is inspired from \cite{mildenhall2022nerf}, which embeds a single RAW RGB image into multiple representations based on the rolling shutter speed. Since rolling shutter speed is not applicable to RAW \ac{TIR} images, we use the temperature periodic constant $D$ to simulate a similar effect.

Varying $D$ controls the appearance of the thermal embedding: smaller $D$ values create sharp edges with artifacts, while larger $D$ values can saturate the image. Selecting $D$ values does not require extensive prior knowledge of the environment, but an approximate range. We find that randomly sampling $D$ values between 4.5 and 45 adapts well to various outdoor environments.
\subsubsection{Intuition behind thermal embedding}
A 14-bit RAW \ac{TIR} pixel value spans a much wider range than 8-bit tone-mapped images, making it difficult to represent small temperature differences. Multichannel thermal embedding uses multiple sinusoidal functions to map temperatures to different intensities, allowing higher resolution representation of small temperature differences. This method simulates \ac{TIR} images from diverse environments and provides a detailed, temperature-aware representation of 14-bit RAW \ac{TIR} images.

\subsection{Adaptive channel compression}

  


A critical aspect of thermal embedding involves determining the optimal temperature period $D$ to ensure effective representation of the given \ac{TIR} image for the downstream task. While employing multiple channels can enrich feature representation, not all multichannel embeddings yield beneficial results; some may introduce artifacts and degrade performance. Thus, relying solely on multichannel thermal embedding requires prior knowledge to estimate the optimal $D$ for the task.

However, adaptive channel compression eliminates the need to estimate $D$ values by applying task-specific tone-mapping weights to each thermal embedding, as shown in \figref{fig:adaptive_overview}. Consider a multichannel thermal embedding,$I_{ME} \in \mathbb{R}^{B \times N \times H \times W}$. With temperature periods sampled from a uniform distribution, each embedding passes through shared CNN with two strided convolutional layers and ReLU activation. The extracted features form a 3-channel feature map, resulting in an output shape of $B \times N \times 3 \times H \times W$. 

Unlike direct channel concatenation in \cite{xu2023toward}, this method applies layer normalization followed by a global average pooling to each channel, yielding $N \times 3$ feature vectors for each thermal embedding. To encode adaptive weights, these vectors are concatenated and passed through \ac{MLP} layers with two 64 hidden units and a single layer with $3N$ layer.  The output, activated by Softmax, forms task-adaptive weights for each thermal embedding. Ultimately, each weighted embedding is tone-mapped into a single 3-channel image through per-channel weighted summation, as shown in Equation \ref{weighted_sum}.

\begin{equation}
    \textbf{X}_{i}^{\prime} = \sum_{n=1}^{N} \mathbf{\omega}_{\text{norm},i}[n] \cdot I_{ME,n}  \quad \text{for } i \in \{1, 2, 3\}
\label{weighted_sum}
\end{equation}
where $\omega$ refers to channel weightings for the compressed channel index $i$ and $I_{ME,n}$ refers to the $n$th thermal embedding. Finally, the compressed input is then passed as a 3-channel input to the downstream task network. 

%% file: experiment.tex
\section{experiments}
\label{sec:experiment}



\begin{figure}[]
  \centering
  \begin{subfigure}{0.3\columnwidth}
    \centering
    \includegraphics[width=\columnwidth]{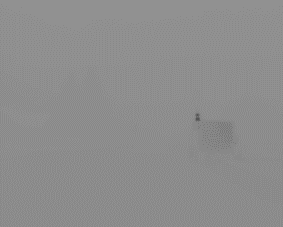}
    \caption{Raw}
    \label{flir_raw}
  \end{subfigure}
  \begin{subfigure}{0.3\columnwidth}
    \centering
    \includegraphics[width=\columnwidth]{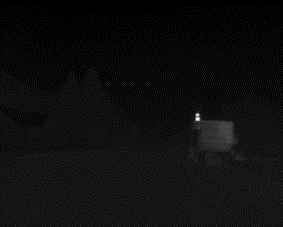}
    \caption{Min-max}
    \label{flir_minmax}
  \end{subfigure}
  \begin{subfigure}{0.3\columnwidth}
    \centering
    \includegraphics[width=\columnwidth]{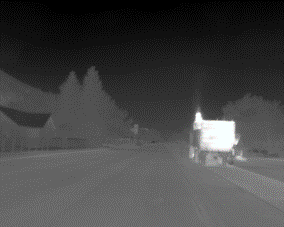}
    \caption{Clip \cite{shin2023wacv}}
    \label{flir_clip}
  \end{subfigure}
  \begin{subfigure}{0.3\columnwidth}
    \centering
    \includegraphics[width=\columnwidth]{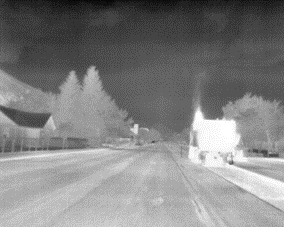}
    \caption{HE \cite{shin2022raliros}}
    \label{flir_HE}
  \end{subfigure}
  \begin{subfigure}{0.3\columnwidth}
    \centering
    \includegraphics[width=\columnwidth]{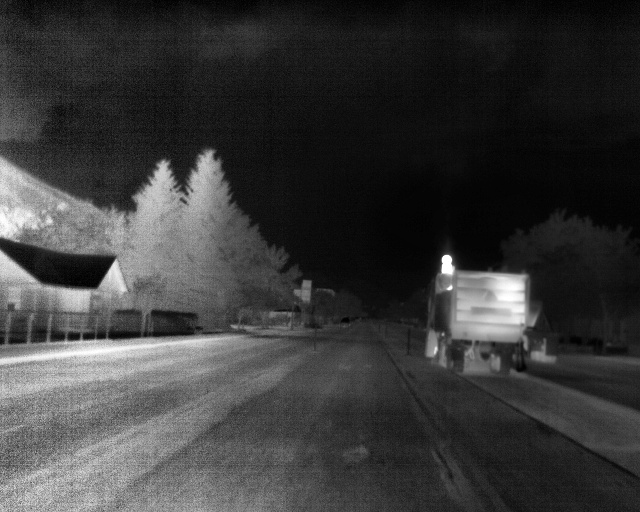}
    \caption{Fieldscale \cite{gil2024fieldscale}}
    \label{flir_fieldscale}
    
  \end{subfigure}
    \begin{subfigure}{0.3\columnwidth}
    \centering
    \includegraphics[width=\columnwidth]{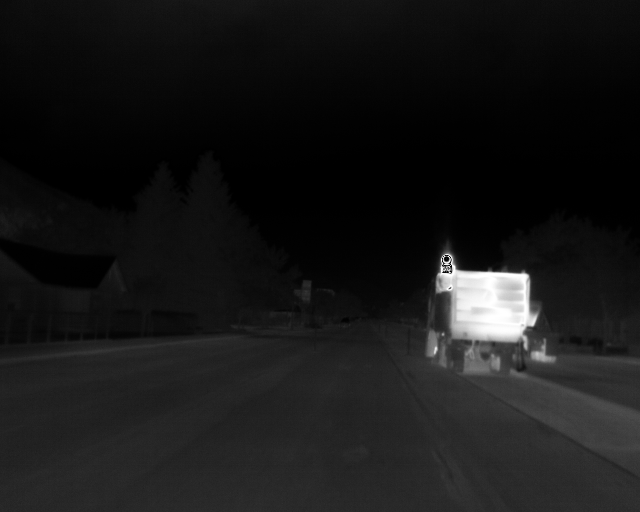}
    \caption{Thermal Embedding}
    \label{flir_te}
    
  \end{subfigure}
  \caption{\bl{Visualization examples of commonly used baseline \ac{TIR} tone-mapping methods. $D$ value of 3 is used for thermal embedding}}
  \label{fig:rescale-visualization}

\end{figure}

\subsection{Experiment Overview}

We evaluate the performance of \ac{TCNet} on two popular downstream tasks: object detection and monocular depth estimation. In addition, we analyze the task adaptive characteristic of \ac{TCNet} based on its intermediate tone-mapped images for each downstream task. 

\subsubsection{Datasets}

\input{tables/dataset-info}

\tabref{dataset-info} presents an overview of the datasets used in our experiments. To use \ac{TCNet}, the \ac{TIR} image must be stored in its RAW format by a radiometric \ac{TIR} camera with known camera offset parameters. As a result, only tasks that fulfill this requirement are object detection and self-supervised depth estimation. For each task, we used the FLIR-ADAS v2 \cite{fliradas} dataset for object detection and VIVID \cite{lee2022vivid++} benchmark for depth estimation. We further assessed zero-shot performance on unseen datasets with significant temperature and geographical variations. For object detection, we manually annotated the valley sequence from the STheReO \cite{yun2022sthereo} dataset based on classes overlapping with the FLIR dataset. For depth estimation, we selected the MS2 dataset \cite{shin2023cvpr} test sequence. For all evaluations, TCNet was trained exclusively on the training set and subsequently tested various test sets.

\subsubsection{Baselines}

We used different 14-bit to 8-bit \ac{TIR} image rescaling methods as the baseline models. Tone-mapped visualizations of the baselines are shown in \figref{fig:rescale-visualization}. Descriptions to the baseline tone-mapping methods are shown below:

\begin{itemize}
    \item RAW: Linear rescaling with $2^{14}-1$ (max resolution)
   
    \item Min-max: Linear rescaling with min/max value 
    
    \item Clip \cite{shin2023wacv}: Locally consistent linear rescaling with 1\% and 99\% of the image intensity used as the clipping parameters.

    \item HE \cite{shin2022raliros}: Bin-based histogram equalization used. Bin size of 30 is used.

    \item Fieldscale \cite{gil2024fieldscale}: 2D min and max field-based non-linear rescaling. Used provided default parameters. 
    \item TE($N$): $N$ channel thermal embedding.

\end{itemize}

\subsection{Evaluation on object detection}


\subsubsection{Implementation Details}



We evaluated three object detection architectures: RetinaNet \cite{lin2017focal}, YOLOX \cite{ge2021yolox}, and Sparse-RCNN \cite{sparsercnn}, using baselines from a previous work \cite{xu2023toward}. No changes were made to the downstream models' architectures, except for \ac{TCNet}, which was integrated into the front end. We uniformly sampled $N$ temperature periods within the 4.5 to 45 range, as periods beyond this range offered no additional benefit.

The detection networks were trained according to their original loss functions, excluding color augmentations and image resizing. Except for Thermal Embedding (TE) and \ac{TCNet}, all baselines used 3-channel \ac{TIR} images by repeating the single-channel image to utilize ImageNet initialization, except for YOLOX. Models were trained on $640 \times 512$ pixel images, and the detection performance was evaluated using mean average precision (mAP) following standard COCO evaluation protocols. 

\subsubsection{Evaluation on FLIR}

\input{tables/obj_det_table_mAP}


\tabref{main_results_table} shows the detection result evaluated on the FLIR-ADAS dataset. As predicted, the direct use of RAW images worsened performance compared to 8-bit rescaled images. In contrast, \ac{TCNet} outperformed all baselines by 12.4\%, 5.2\%, and 7.8\% for RetinaNet, YOLOX, and Sparse-RCNN respectively. Even using thermal embedding alone (TE(3)) improved performance over the baselines. Adaptive channel compression further enhanced detection performance by 6.1\%, 4.9\%, and 4.1\% for RetinaNet, YOLOX, and Sparse-RCNN respectively. However, increasing the number of thermal embeddings without adaptive channel compression reduced performance, likely because detection networks are optimized for 3-channel RGB images.


\subsubsection{\bl{Zero-shot Evaluation on STheReO}}

\tabref{main_results_table_valley} shows the zero-shot detection results on the STheReO dataset. TCNet also demonstrates the best overall performance over the baseline models, achieving either the best or second best performance in all metrics. We posit such improvement originates from the use of thermal embedding, effectively simulating a temperature-aware data augmentation. This is critical because thermal image datasets are typically small, and they capture a limited range of environments. As a solution, TCNet's thermal embedding diversifies the representation of \ac{TIR} images, mimicking different environmental temperatures during training. This approach enhances the model's robustness to environmental changes, making TCNet particularly effective in generalizing to unseen data. This compensates for the limited diversity within thermal image datasets and ensuring reliable object detection across diverse temperature conditions.

For advanced models like YOLOX that use extensive geometric augmentation, the performance gap between \ac{TCNet} and linear scaling methods decreases on the STheReO dataset. This is likely due to the small temperature and scene variations in the test set and the small evaluation size, resulting in fewer image degradation issues. Nonetheless, \ac{TCNet} combined with geometric augmentation still yields the best detection performance.

\subsection{Evaluation on geometric tasks}

\subsubsection{Implementation Details}

We evaluated \ac{TCNet} on two geometric tasks: monocular depth estimation and pose estimation. As for the baseline model, we selected Monodepth2-based \ac{TIR} depth estimation model from prior research \cite{shin2022raliros}, which is a self-supervised model for estimating both depth and pose. During training, we followed the original hyperparameters and loss functions outlined by \cite{shin2022raliros}, with an exception to the training epochs which we extended it to 400. All models leveraged ImageNet preinitialization on its backbone. 
Additionally, we also compare it with multi-spectral \cite{shin2021ralcolor,shin2023wacv} models. For evaluation on the MS2 dataset, we only evaluate models whom have open-sourced training code or pre-trained model available. Given that both depth and pose model use a shared encoder, we also use a shared \ac{TCNet} for both task, training \ac{TCNet} in a multi-task manner.



\subsubsection{Experiment results}
\input{tables/indoor_depth_results}

\bl{\tabref{tab:depth_outdoor} and \tabref{ms2_depth} presents monocular depth estimation results evaluated on VIVID \cite{lee2022vivid++} as well as zero-shot evaluation on the MS2 dataset \cite{shin2023cvpr} respectively}. \ac{TCNet} surpasses not only models trained with existing baselines, but also outperforms models trained with multi-spectral data; it especially shows strong generalization performance to unseen environment. Aligning with the previous findings, the use of thermal embedding on the MS2 dataset \cite{shin2023cvpr} improves the performance, showcasing its robustness to unseen data despite the difference in the average temperature and type of the scene. Out of the conventional methods, we find that HE and Clip yields the best depth estimation performance on the VIVID and MS2 dataset respectively. Thus, without extensive prior knowledge of the environment, non-optimal performance is achieved. On the contrary, with task-adaptive tone-mapping element, \ac{TCNet}, is able to achieve consistent performance across diverse environments. For ego-pose estimation, which is the other counterpart, we observe a similar trend; results related to this is further discussed in the supplementary material. 



\subsection{Task-adaptivity Analysis}

\subsubsection{Quantitative evaluation}

Image entropy can be considered as one of the key qualitative metric for determining the contrast of an image. Similarly, high level image characteristics are evaluated by measuring the perceptual similarity of two images using pretrained backbones; notably recent methods has shifted towards using multimodal embedding models trained on large corpus of data such as CLIP \cite{radford2021clip}. 

To verify our hypothesis on task-adaptive nature of \ac{TCNet}, we examine both low and high level image characteristics of tone mapped images. For low level characteristic, we compute the image entropy of the tone mapped image per each task. For high level characteristic, we adopt coarsely paired RGB-\ac{TIR} evaluation sets from FLIR, STheReO, and VIVID test set to compute the perceptual similarity between coarse image pairs. For this, we leverage ImageBind \cite{girdhar2023imagebind}, a large multimodal feature embedding model that supports both RGB and \ac{TIR} images to compute embedding vector for \bl{each daylight RGB and \ac{TIR} image pairs and measure their cosine similarity.}

\input{tables/task_adaptivity_table}

\begin{figure*}[t]
  \centering
  \begin{subfigure}{0.3\textwidth}
    \centering
    \includegraphics[width=\textwidth]{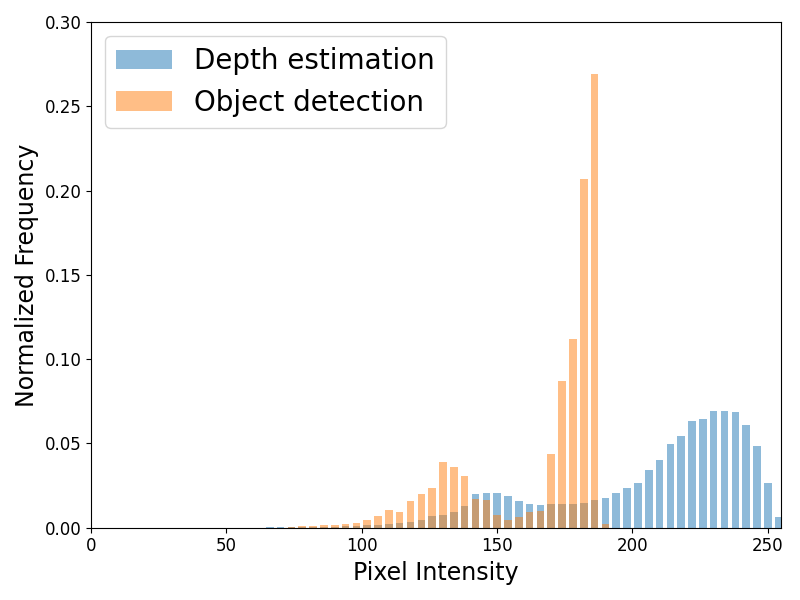}
    \caption{FLIR}
    \label{flir_distribution}
  \end{subfigure}
  \begin{subfigure}{0.3\textwidth}
    \centering
    \includegraphics[width=\textwidth]{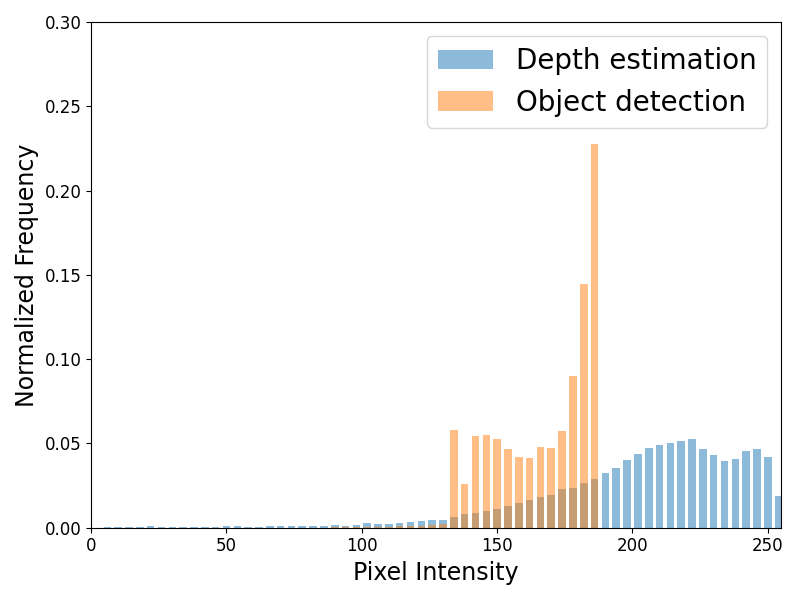}
    \caption{STheReO Valley }
    \label{valley_distribution}
  \end{subfigure}
  \begin{subfigure}{0.3\textwidth}
    \centering
    \includegraphics[width=\textwidth]{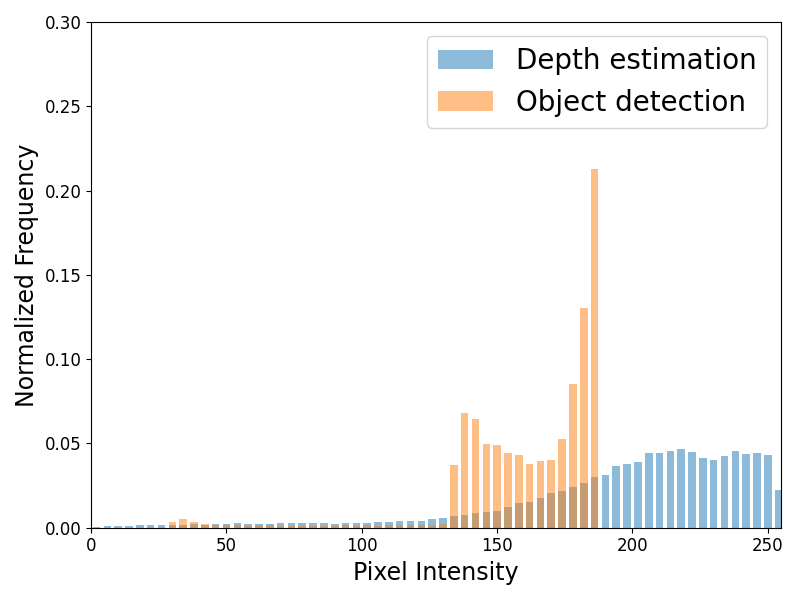}
    \caption{VIVID }
    \label{vivid_distribution}
  \end{subfigure}
  \caption{Average image histogram distribution of the tone-mapped images across different datasets. Different tasks tone-map the same \ac{TIR} images differently.}
  \label{tone-map-divergence}
  \vspace{-5mm}
\end{figure*}

\tabref{task-adaptivity-tonemap} shows the characteristics of tone-mapped images for object detection and monocular depth estimation. We observe that tone-mapped images for depth estimation tend to have higher image entropy compared to those for object detection, which yield higher perceptual similarity. For depth estimation, higher entropy means higher contrast in the low-level pixel space, which is crucial for sharp depth near object boundaries. In contrast, object detection relies on feature vectors extracted from deeper layers, requiring high semantic validity and resulting in high perceptual similarity to RGB images despite lower contrast. \bl{Notably, the STheReO dataset shows negative perceptual similarity, unlike the other datasets. This result likely stems from the need to resize RGB images for use with the ImageBind.}

To further demonstrate the task adaptivity, we visualize the average data distribution of the tone-mapped images from each dataset in \figref{tone-map-divergence}. Here, we identify that each task exhibit different tone-mapped distribution, with differing KL-divergence values; thus, demonstrating that \ac{TCNet} tonemaps different \ac{TIR} images per different task. 


\begin{figure}
  \centering 
\includegraphics[width=\columnwidth]{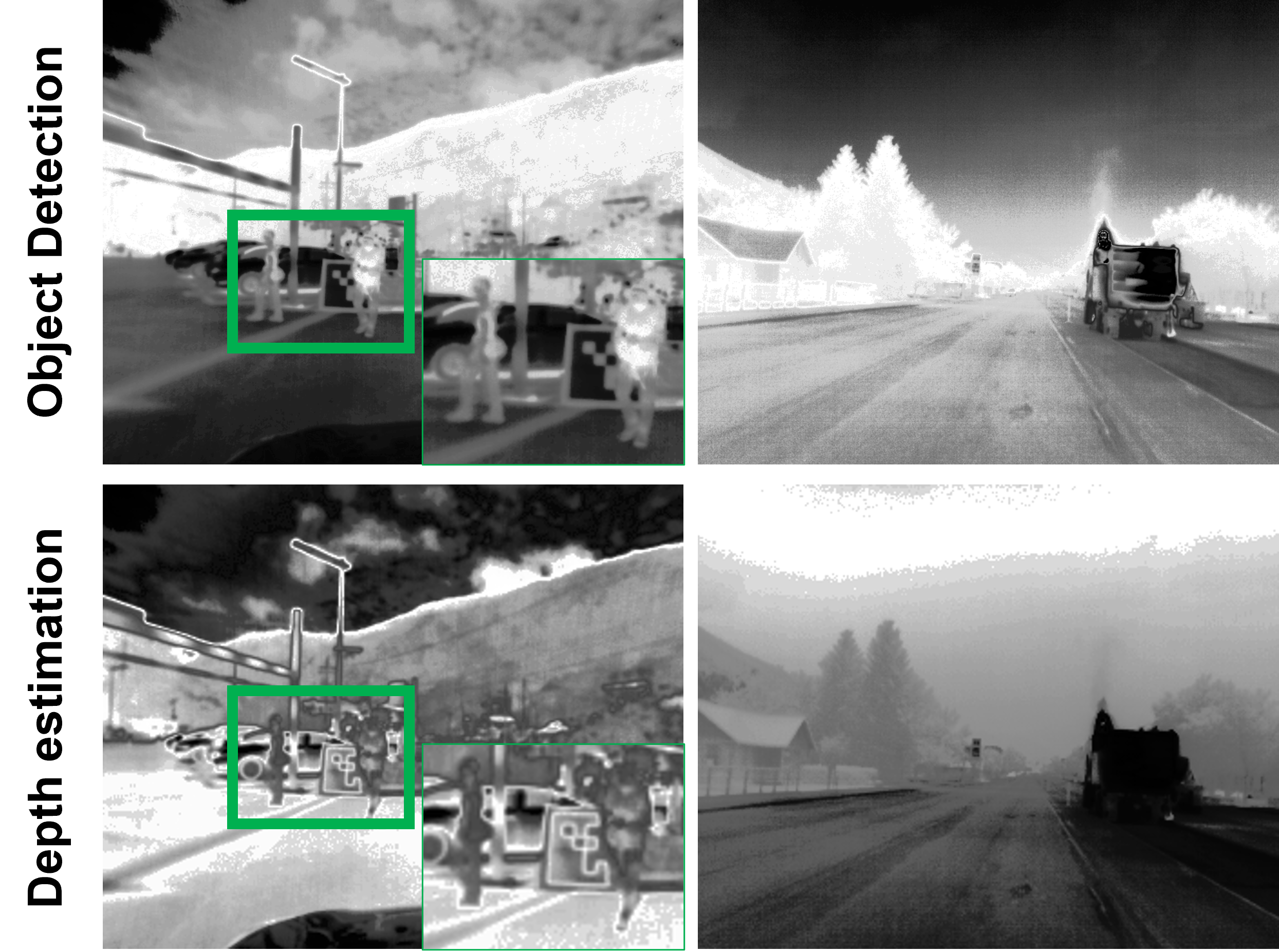}
\caption{Example tone-mapped \ac{TIR} images for object detection and depth estimation. Object detection focuses on preserving object specific characteristics while depth focuses on preserving sharp boundaries between background and objects and neglecting dynamic objects within the scene. See supplementary for more visualizations.}
\label{fig:task-adaptive-tonemap}

\end{figure}

\subsubsection{Qualitative evaluation}


\figref{fig:task-adaptive-tonemap} illustrates the tone-mapped visualizations of \ac{TCNet} for object detection and depth estimation. In both tasks, we observe different characteristics highlighted for each image. For object detection, it focuses on highlighting the thermal characteristics of the objects of interest (green box) while it neglects fine details of the background. Conversely, for depth estimation, it tends to neglect details on the objects within the image, but rather focuses on achieving finer edges and boundaries in the background, which is pivotal for depth estimation. 

\subsubsection{Artifact rejection}


\begin{figure}[]
  \centering
  \begin{subfigure}{0.3\columnwidth}
    \centering
    \includegraphics[width=\columnwidth]{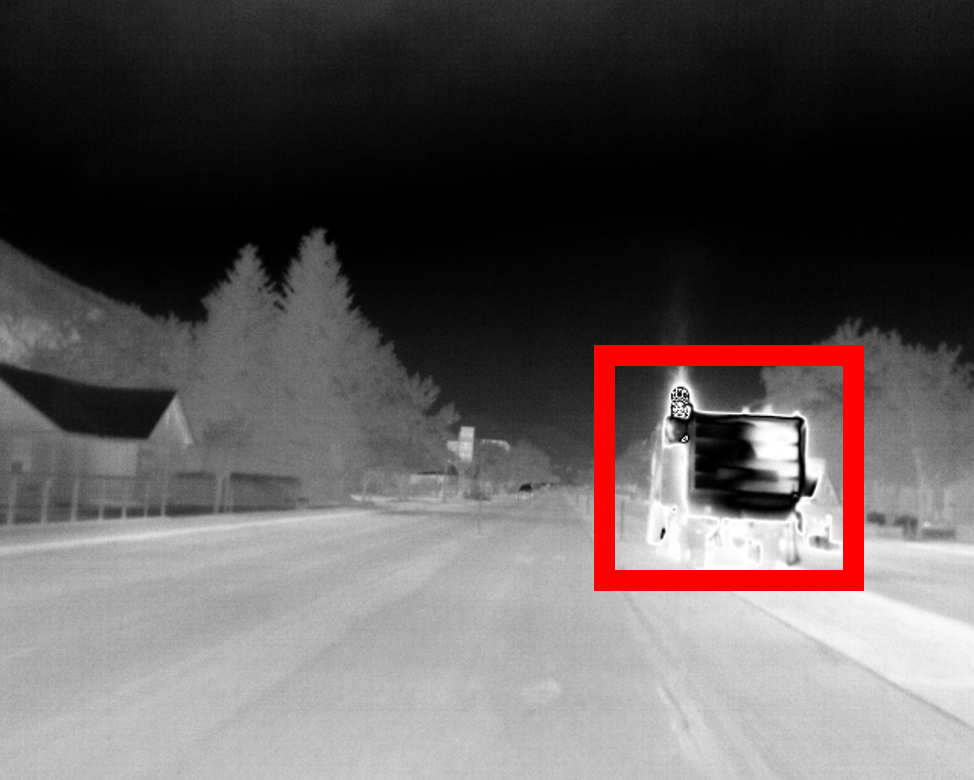}
    \caption{$D = 4.5$}
    \label{te1}
  \end{subfigure}
  \begin{subfigure}{0.3\columnwidth}
    \centering
    \includegraphics[width=\columnwidth]{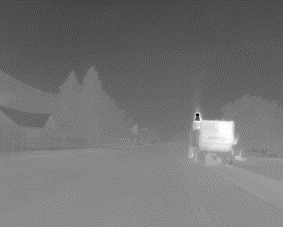}
    \caption{$D = 13.5$}
    \label{te2}
  \end{subfigure}
  \begin{subfigure}{0.3\columnwidth}
    \centering
    \includegraphics[width=\columnwidth]{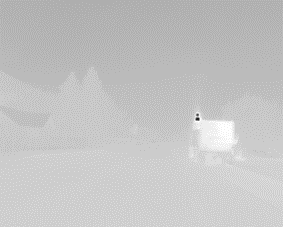}
    \caption{$D = 45.0$}
    \label{te3}
  \end{subfigure}
  \caption{Thermal embeddings with their respective thermal temperature periods.}
  \label{tone-map-artifact}

\end{figure}

\input{tables/task-adaptive-weights}

To verify artifact rejection capability through adaptive channel compression, we analyze the weights assigned for each downstream task. \figref{tone-map-artifact} illustrates the sample thermal embedding used for the task-adaptive tone-mapping, and \tabref{task-adaptive-weights} shows their respective weightings assigned for each image in the last layer of the adaptive channel compression. Here, we observe a presence of image artifacts in the first thermal embedding (red box). Upon the presence of such artifacts, we observe that the adaptive channel compression assign lowest weightings for the first thermal embedding for both tasks. 
However, what we find interesting is that images tone-mapped with object detection tends to be more lenient towards image artifacts than depth estimation, yielding images that retain thermal characteristic of the object. In contrast, depth estimation favors images that highlight the background details while suppressing the details of the objects (green box) that might result in depth uncertainties. 

\subsection{Ablation studies}

\subsubsection{Influence on number of channels}

\input{tables/obj_det_table_channel}

\input{tables/indoor_depth_channel}

\tabref{tab:channel-flir} and \tabref{tab:depth_outdoor_channels} show the change in performance when varying the number of channels. In object detection, increasing the number of channels improves performance, likely due to the network encountering more diverse representations of TIR images. This suggests that for semantic tasks like object detection, additional channels enhance the model's robustness by introducing more variability in object appearances.


Conversely, for depth estimation, the best results are achieved with the original channel configuration. Although adding more channels reduces performance, the outcomes remain competitive with baseline models. This reduction in performance may stem from the decreased photometric consistency between consecutive frames that accompanies a higher number of channels, highlighting a task-specific trade-off in channel optimization.



\input{tables/detection-sensitivity.tex}

\subsubsection{Sensitivity to random period values}

A crucial aspect of TCNet's configuration is the selection of the temperature period $D$. While random sampling of $D$ during training is viable, it raise concerns about potentially adverse effects on the performance during inference. To address this, we assessed the impact of random sampling by conducting ten iterations with randomly predefined seeds on TCNet and thermal embeddings trained with YOLOX, as shown in \tabref{tab:random_sensitivity}. Interestingly, a higher number of channels yielded the lowest standard deviation $10^{-4}$ across iterations, showing reduced variability in performance. This suggests that randomness in $D$ sampling does not detrimentally affect performance, but it rather mitigates the stochasticity associated with random $D$ values, further demonstrating its efficacy in enhancing TCNet's performance consistency.



\input{tables/flops_table}

\subsubsection{Computational Cost}

An important consideration when using \ac{TCNet} is the increased computational cost. We analyzed the computational cost of Sparse-RCNN, the heaviest model, by measuring the \ac{FLOPs}, as shown in \tabref{tab:flops}. \bl{We denoted the runtime for thermal embedding measured on Intel i9-12900 CPU and noted the additional \ac{FLOPs} incurred due to either adaptive channel compression or modifications to the convolutional backbone. Our analysis shows that increasing the number of channels for thermal embedding results in a linear rise in computational cost. However, employing adaptive channel compression adds only 0.1 GFLOPs while providing an 8.2\% gain in mAP. In contrast, using thermal embedding without adaptive compression adds 1.8 GFLOPs. Thus, adaptive channel compression not only improves performance but also is more efficient than solely using thermal embedding.}

  
\subsubsection{Comparison with other color augmentations}
\input{tables/color-aug}

As shown in \tabref{color-augmentation}, we compared thermal embedding with other commonly used color augmentations. For this, we trained Sparse-RCNN with augmentations applied to min-max \ac{TIR} images. As the results demonstrate, thermal embedding induces performance improvement from simulating \ac{TIR} images captured at diverse environments, surpasses other color augmentation methods. With adaptive channel compression (\ac{TCNet}), we observe a further improvement in the performance.




%% file: tables/dataset-info.tex
\begin{table}[]
\centering
\caption{Dataset overview}
\label{dataset-info}
\resizebox{\columnwidth}{!}{%
\begin{tabular}{cccccc}
\hline \hline
Dataset                                                                & \begin{tabular}[c]{@{}c@{}}Mean\\ Temperature\\ \degreecelsius\end{tabular} & \begin{tabular}[c]{@{}c@{}}S.t.d\\ Temperature\\ \degreecelsius\end{tabular} & Location                                                                     & Scene                                                                 & \# of images \\ \hline
FLIR-ADAS v2 - Train                                                   & 18.00                                                                                      & 14.30                                                                                       & \begin{tabular}[c]{@{}c@{}}USA (California), \\ England, France\end{tabular} & Urban                                                                 & 10,724       \\
FLIR-ADAS v2 - Test                                                    & 18.49                                                                                      & 15.04                                                                                       & \begin{tabular}[c]{@{}c@{}}USA: California, \\ Idaho, Michigan\end{tabular}  & \begin{tabular}[c]{@{}c@{}}Urban,\\ Sub-urban\end{tabular}            & 2,151        \\
\begin{tabular}[c]{@{}c@{}}STheReO - Valley\\ (Test)\end{tabular}      & 26.34                                                                                      & 8.12                                                                                        & Korea: Daejeon                                                               & Sub-urban                                                             & 955          \\
\begin{tabular}[c]{@{}c@{}}VIVID - Outdoor Day\\ (Train)\end{tabular}  & 15.03                                                                                      & 12.67                                                                                      & Korea: Daejeon                                                               & Campus                                                                & 2,213        \\
\begin{tabular}[c]{@{}c@{}}VIVID - Outdoor Night\\ (Test)\end{tabular} & 15.57                                                                                      & 7.90                                                                                        & Korea: Daejeon                                                               & Campus                                                                & 2,019        \\
MS2 - Test                                                             & 30.70                                                                                      & 2.77                                                                                        & Korea: Daejeon                                                               & \begin{tabular}[c]{@{}c@{}}Campus, \\ Urban, \\ Resident\end{tabular} & 22,919       \\ \hline \hline
\end{tabular}%
}
\end{table}

%% file: tables/obj_det_table_mAP.tex
\begin{table}[]
\caption{Quantitative detection results on FLIR ADAS \cite{fliradas}. Best results highlighted in \textbf{bold}; Second best in \underline{underlines}.}
\label{main_results_table}
\resizebox{\columnwidth}{!}{%
\begin{tabular}{c|ccccccc}
\hline \hline
Dataset                      & \multicolumn{7}{c}{FLIR ADAS}                                                                                                                                                                                                                                                                                \\ \hline \hline
Model                        & Baseline   & mAP           & \begin{tabular}[c]{@{}c@{}}mAP\\ @0.5\end{tabular} & \begin{tabular}[c]{@{}c@{}}mAP\\ @0.75\end{tabular} & \begin{tabular}[c]{@{}c@{}}mAP\\ @small\end{tabular} & \begin{tabular}[c]{@{}c@{}}mAP\\ @medium\end{tabular} & \begin{tabular}[c]{@{}c@{}}mAP\\ @large\end{tabular} \\ \hline 
\multirow{9}{*}{RetinaNet}   & FLIR AGC   & 25.2          & 49.8                                               & 22.5                                                & 14.7                                                  & 49.4                                                  & 58.4                                                 \\
                             & Clip \cite{shin2023wacv}      & 24.1          & 45.8                                               & 22.9                                                & 12.1                                                  & 50.9                                                  & 60.0                                                 \\
                             & HE \cite{shin2022raliros}         & 23.4          & 46.0                                               & 21.0                                                & 11.8                                                  & 48.9                                                  & 60.6                                                 \\
                             & Min-max    & 22.3          & 49.2                                               & 15.9                                                & 12.7                                                  & 44.5                                                  & 51.6                                                 \\
                             & RAW        & 20.1          & 37.4                                               & 19.1                                                & 6.1                                                   & 45.8                                                  & 58.5                                                 \\
                             & FieldScale \cite{gil2024fieldscale} & 20.6          & 49.8                                               & 23.9                                                & 14.5                                                  & 52.3                                                  & 57.7                                                 \\ \cline{2-8} 
                             & TE(3)      & \underline{29.1}          & \underline{50.8}                                               & \underline{28.7}                                                & \underline{15.9}                                                  & \underline{59.4}                                                  & \underline{64.2}                                        \\
                             & TE(10)     & 28.6          & 50.0                                               & 27.8                                                & 15.4                                                  & 57.1                                                  & \underline{64.2}                                        \\
                             & TCNet      & \textbf{34.7} & \textbf{53.3}                                      & \textbf{37.4}                                       & \textbf{22.7}                                         & \textbf{64.1}                                         & \textbf{67.7}                                                 \\ \hline
\multirow{9}{*}{YOLO-X}      & FLIR AGC   & 53.5          & 81.4                                               & 57.6                                                & \underline{45.8}                                                  & 70.0                                                  & 71.5                                                 \\
                             & Clip \cite{shin2023wacv}      & 49.1          & 75.4                                               & 52.9                                                & 42.1                                                  & 65.6                                                  & 58.9                                                 \\
                             & HE \cite{shin2022raliros}        & 40.7          & 64.9                                               & 41.8                                                & 31.4                                                  & 63.4                                                  & 64.5                                                 \\
                             & Min-max    & 50.8          & 81.6                                               & 53.9                                                & 43.1                                                  & 67.5                                                  & 71.7                                                 \\
                             & RAW        & 37.9          & 63.3                                               & 39.5                                                & 27.1                                                  & 62.2                                                  & 62.1                                                 \\
                             & FieldScale \cite{gil2024fieldscale} & \underline{54.0}          & \underline{82.2}                                               & \underline{59.8}                                                & \textbf{47.8}                                         & \underline{73.0}                                                  & \underline{75.8}                                                 \\ \cline{2-8} 
                             & TE(3)      & 53.3          & 80.9                                               & 58.2                                                & 44.8                                                  & 72.0                                                  & 73.9                                                 \\
                             & TE(10)     & 51.1          & 78.4                                               & 54.8                                                & 41.5                                                  & 72.1                                                  & 76.9                                                 \\
                             & TCNet      & \textbf{56.0} & \textbf{83.8}                                      & \textbf{60.5}                                       & \textbf{47.8}                                         & \textbf{73.4}                                         & \textbf{77.0}                                        \\ \hline
\multirow{9}{*}{Sparse-RCNN} & FLIR AGC   & 37.5          & 60.9                                               & 41.0                                                & 29.2                                                  & 55.0                                                  & 63.4                                                 \\
                             & Clip \cite{shin2023wacv}       & 38.4          & 61.8                                               & 42.4                                                & 30.6                                                  & 56.7                                                  & 62.4                                                 \\
                             & HE \cite{shin2022raliros}         & 39.1          & 62.5                                               & 43.1                                                & 32.2                                                  & 54.8                                                  & 61.8                                                 \\
                             & Min-max    & 37.9          & 63.0                                               & 39.7                                                & 31.1                                                  & 52.8                                                  & 58.6                                                 \\
                             & RAW        & 12.3          & 24.9                                               & 11.7                                                & 7.9                                                   & 22.8                                                  & 19.8                                                 \\
                             & FieldScale \cite{gil2024fieldscale} & 37.2          & 59.3                                               & 40.6                                                & 27.9                                                  & 56.6                                                  & \underline{66.0}                                        \\ \cline{2-8} 
                             & TE(3)      & \underline{41.6}          & \underline{65.3}                                               & \underline{45.0}                                                & \underline{32.5}                                                  & \underline{61.1}                                                  & \underline{63.7}                                                 \\
                             & TE(10)     & 26.8          & 43.9                                               & 29.2                                                & 19.9                                                  & 43.3                                                  & 33.2                                                 \\
                             & TCNet      & \textbf{45.7} & \textbf{70.5}                                      & \textbf{49.8}                                       & \textbf{36.1}                                         & \textbf{67.5}                                         & \textbf{69.1}                                                 \\ \hline \hline
\end{tabular}%
}
\end{table}

\begin{table}[]
\caption{Quantitative detection results on STheReO \cite{yun2022sthereo}. Best results highlighted in \textbf{bold}; Second best in \underline{underlines}.}
\label{main_results_table_valley}
\resizebox{\columnwidth}{!}{%
\begin{tabular}{c|ccccccc}
\hline \hline
Dataset                      & \multicolumn{7}{c}{STheReO - Valley}                                                                                                                                                                                                                                                                        \\ \hline \hline
Model                        & Baseline   & mAP           & \begin{tabular}[c]{@{}c@{}}mAP\\ @0.5\end{tabular} & \begin{tabular}[c]{@{}c@{}}mAP\\ @0.75\end{tabular} & \begin{tabular}[c]{@{}c@{}}mAP\\ @small\end{tabular} & \begin{tabular}[c]{@{}c@{}}mAP\\ @medium\end{tabular} & \begin{tabular}[c]{@{}c@{}}mAP\\ @large\end{tabular} \\ \hline
\multirow{9}{*}{RetinaNet}   & FLIR AGC   & -             & -                                                  & -                                                   & -                                                    & -                                                     & -                                                    \\
                             & Clip       & 31.2          & 57.4                                               & 29.9                                                & 7.9                                                  & 40.6                                                  & 59.9                                                 \\
                             & HE         & 29.6          & 56.8                                               & 26.5                                                & 6.9                                                  & 38.7                                                  & 58.1                                                 \\
                             & Min-max    & 20.8          & 35.6                                               & 21.5                                                & 5.0                                                    & 21.2                                                  & 30.2                                                 \\
                             & RAW        & 27.7          & 54.9                                               & 24.4                                                & 7.9                                                  & 35.4                                                  & 52.4                                                 \\
                             & FieldScale & 21.9          & 46.7                                               & 18.6                                                & 5.0                                                  & 28.1                                                  & 49.6                                                 \\ \cline{2-8} 
                             & TE(3)      & \underline{35.4}          & \textbf{62.2}                                      & \underline{35.9}                                                & 10.0                                                   & \underline{45.8}                                                  & \underline{62.7}                                                 \\
                             & TE(10)     & 33.2          & \underline{61.4}                                               & 32.5                                                & \underline{10.3}                                                 & 42.5                                                  & 57.5                                                 \\
                             & TCNet      & \textbf{36.1} & 62.0                                                 & \textbf{37.3}                                       & \textbf{12.6}                                        & \textbf{46.8}                                         & \textbf{66.6}                                        \\ \hline
\multirow{9}{*}{YOLO-X}      & FLIR AGC   & -             & -                                                  & -                                                   & -                                                    & -                                                     & -                                                    \\
                             & Clip       & 47.0            & 80.6                                               & 47.7                                                & 24.7                                                 & 56.2                                                  & 73.8                                                 \\
                             & HE         & 45.3          & 76.8                                               & 45.5                                                & 22.3                                                 & 54.1                                                  & 72.8                                                 \\
                             & Min-max    & 46.5          & \textbf{82.0}                                      & 46.2                                                & \textbf{25.5}                                        & 55.1                                                  & 69.9                                                 \\
                             & RAW        & 40.9          & 70.4                                               & 41.1                                                & 17.0                                                   & 49.6                                                  & 64.1                                                 \\
                             & FieldScale & 47.1          & 79.2                                               & 48.4                                                & 22.7                                                 & \underline{57.1}                                                  & \textbf{76.8}                                                 \\ \cline{2-8} 
                             & TE(3)      & 46.8          & 80.4                                               & 47.9                                                & 24.3                                                 & 55.8                                                  & 73.4                                                 \\
                             & TE(10)     & \underline{47.5}          & 80.5                                               & \underline{49.0}                                                  & 24.3                                                 & 56.5                                                  & 75.1                                                 \\
                             & TCNet      & \textbf{48.3} & \underline{81.3}                                                 & \textbf{50.0}                                         & \underline{24.9}                                                 & \textbf{57.6}                                         & \underline{76.4}                                        \\ \hline
\multirow{9}{*}{Sparse-RCNN} & FLIR AGC   & -             & -                                                  & -                                                   & -                                                    & -                                                     & -                                                    \\
                             & Clip       & 22.7          & 45.4                                               & 19.8                                                & 10.5                                                 & 30.3                                                  & 32.1                                                 \\
                             & HE         & 10.7          & 20                                                 & 10.1                                                & 2.1                                                  & 14.6                                                  & 31.1                                                 \\
                             & Min-max    & 34.9          & 64                                                 & 34.5                                                & 15.6                                                 & 42.4                                                  & 57.9                                                 \\
                             & RAW        & 4.1           & 9.5                                                & 3.7                                                 & 0.8                                                  & 5.9                                                   & 6.5                                                  \\
                             & FieldScale & 31.3          & 58.3                                               & 30.8                                                & 13.0                                                 & 41.0                                                  & 47.3                                                 \\ \cline{2-8} 
                             & TE(3)      & \underline{38.3}          & \underline{71.2}                                               & \underline{37.7}                                                & \underline{17.3}                                                 & \underline{48.5}                                                  & \textbf{69.1}                                        \\
                             & TE(10)     & 17.9          & 37.8                                               & 14.1                                                & 9.9                                                  & 23.0                                                  & 20.0                                                 \\
                             & TCNet      & \textbf{39.5} & \textbf{73.2}                                      & \textbf{38.2}                                       & \textbf{18.8}                                        & \textbf{49.4}                                         & \underline{63.5}                                                 \\ \hline \hline
\end{tabular}%
}
\end{table}

%% file: tables/indoor_depth_results.tex
\begin{table}[]
\centering
\caption{Depth estimation result on VIVID. Best results highlighted in \textbf{bold}; Second best in \underline{underlines}.}
\label{tab:depth_outdoor}
\resizebox{\columnwidth}{!}{%
\begin{tabular}{c|cccc|ccc}
\hline \hline
\multirow{2}{*}{Method} & \multicolumn{4}{c|}{Error}                                        & \multicolumn{3}{c}{Accuracy}                                                                                                                                                           \\ \cline{2-8} 
                        & AbsRel         & SqRel          & RMS            & RMSlog         & \begin{tabular}[c]{@{}c@{}}$ \delta$\\ 1.25\end{tabular} & \begin{tabular}[c]{@{}c@{}}$ \delta$\\ 1.25$^2$\end{tabular} & \begin{tabular}[c]{@{}c@{}}$ \delta$\\ 1.25$^3$\end{tabular} \\ \hline
Raw                     & 0.633          & 9.006         & 11.454         & 0.597          & 0.361                                                    & 0.572                                                        & 0.717                                                        \\
Minmax                  & 0.552          & 7.875          & 10.616         & 0.539          & 0.440                                                    & 0.633                                                        & 0.767                                                        \\
HE                      & 0.109          & \underline{0.703}          & \underline{4.132}          & \underline{0.150}           & 0.887                                                    & 0.98                                                         & \underline{0.994}                                                        \\
Clip                    & 0.132          & 0.926          & 5.090           & 0.182          & 0.823                                                    & 0.965                                                        & 0.99                                                         \\
Fieldscale              & 0.544          & 8.158          & 10.874         & 0.532          & 0.453                                                    & 0.643                                                        & 0.769                                                        \\ \hline
Shin \cite{shin2021ralcolor}                   & 0.157          & 1.179          & 5.802          & 0.211          & 0.750                                                    & 0.948                                                        & 0.985                                                        \\
Shin \cite{shin2023wacv}                   & 0.111          & 0.778          & 4.177          & 0.153          & 0.889                                                    & 0.981                                                        & \underline{0.994}                                                        \\ \hline
TE(3)                     & \underline{0.102}          & 0.824          & 4.198          & \underline{0.150}           & \underline{0.904}                                                    & \underline{0.984}                                                        & \underline{0.994}                                                        \\
TE(10)                    & 0.124          & 0.863          & 4.464          & 0.169          & 0.851                                                    & 0.966                                                        & 0.989                                                        \\
TCNet                    & \textbf{0.098} & \textbf{0.611} & \textbf{3.683} & \textbf{0.136} & \textbf{0.915}                                           & \textbf{0.985}                                               & \textbf{0.995}                                               \\ \hline \hline
\end{tabular}%
}
\end{table}

\begin{table}[]
\centering
\caption{Depth results for MS2 dataset. Best results highlighted in \textbf{bold}; Second best in \underline{underlines}.}
\label{ms2_depth}
\resizebox{\columnwidth}{!}{%
\begin{tabular}{c|cccc|ccc}
\hline \hline
\multirow{2}{*}{Method} & \multicolumn{4}{c|}{Error}                                        & \multicolumn{3}{c}{Accuracy}                                                                                                                                                           \\ \cline{2-8} 
                        & AbsRel         & SqRel          & RMS            & RMSlog         & \begin{tabular}[c]{@{}c@{}}$ \delta$\\ 1.25\end{tabular} & \begin{tabular}[c]{@{}c@{}}$ \delta$\\ 1.25$^2$\end{tabular} & \begin{tabular}[c]{@{}c@{}}$ \delta$\\ 1.25$^3$\end{tabular} \\ \hline
Raw                     & 0.477          & 4.473          & 11.174         & 0.654          & 0.278                                                    & 0.514                                                        & 0.72                                                         \\
Minmax                  & 0.784          & 28.167         & 17.554         & 0.742          & \underline{0.384}                                                    & 0.590                                                        & 0.704                                                        \\
HE                      & 0.603          & 9.730          & 12.322         & 0.641          & 0.321                                                    & 0.619                                                        & 0.743                                                        \\
Clip                    & 0.395          & 4.589          & 10.386         & 0.584          & 0.322                                                    & 0.619                                                        & 0.796                                                        \\
Fieldscale              & 0.548          & 6.142          & 13.565         & 0.72           & 0.225                                                    & 0.445                                                        & 0.661                                                        \\ \hline
Shin \cite{shin2021ralcolor}                   & 0.577          & 11.598         & 15.031         & 0.614          & 0.265                                                    & 0.53                                                         & 0.757                                                        \\ \hline
TE(3)                     & 0.447          & 11.442         & 10.960         & \underline{0.467}          & 0.298                                                    & 0.556                                                        & 0.754                                                        \\
TE(10)                    & \underline{0.342}          & \textbf{3.616} & \underline{10.043}         & 0.488          & 0.329                                                    & \underline{0.636}                                                        & \underline{0.836}                                                        \\
TCNet                    & \textbf{0.335} & \underline{4.267}          & \textbf{9.286} & \textbf{0.462} & \textbf{0.435}                                           & \textbf{0.686}                                               & \textbf{0.837}                                               \\ \hline \hline
\end{tabular}%

}
\end{table}

%% file: tables/task_adaptivity_table.tex
\begin{table}[]
\caption{Task-adaptivity analysis for tone-mapped images}
\label{task-adaptivity-tonemap}
\resizebox{\columnwidth}{!}{%
\begin{tabular}{c|cccc|c}
\hline \hline
\multirow{2}{*}{Dataset} & \multicolumn{2}{c}{\begin{tabular}[c]{@{}c@{}}Object\\ Detection\end{tabular}}                                            & \multicolumn{2}{c|}{\begin{tabular}[c]{@{}c@{}}Depth\\ Estimation\end{tabular}}                                           & \begin{tabular}[c]{@{}c@{}}Tone-mapped\\ Distribution\end{tabular} \\ \cline{2-6} 
                         & \begin{tabular}[c]{@{}c@{}}Image\\ Entropy\end{tabular} & \begin{tabular}[c]{@{}c@{}}Perceptual\\ Similarity\end{tabular} & \begin{tabular}[c]{@{}c@{}}Image\\ Entropy\end{tabular} & \begin{tabular}[c]{@{}c@{}}Perceptual\\ Similarity\end{tabular} & KL-Divergence                                                      \\ \hline
FLIR-ADAS                & 2.64                                                    & 0.75                                                            & 3.37                                                    & 0.72                                                            & 13.70                                                              \\
ViVID++                  & 3.73                                                    & 0.72                                                            & 4.03                                                    & 0.69                                                            & 12.68                                                              \\
STheReO                  & 3.72                                                    & -0.40                                                           & 3.96                                                    & -0.48                                                           & 15.01                                                              \\ \hline \hline 
\end{tabular}%
}
\vspace{-5mm}
\end{table}

%% file: tables/task-adaptive-weights.tex
\begin{table}[]
\centering
\caption{Task adaptive weight visualization}
\label{task-adaptive-weights}
\resizebox{\columnwidth}{!}{%
\begin{tabular}{c|cccccc}
\hline \hline
\multirow{3}{*}{Channels}    & \multicolumn{6}{c}{Thermal Embeddings}                                                                                                                                                                                                        \\ \cline{2-7} 
                             & \multicolumn{2}{c}{\begin{tabular}[c]{@{}c@{}}$D = 4.5$\end{tabular}}   & \multicolumn{2}{c}{\begin{tabular}[c]{@{}c@{}}$D = 13.5$\end{tabular}}  & \multicolumn{2}{c}{\begin{tabular}[c]{@{}c@{}}$D = 45.0$\end{tabular}}   \\ \cline{2-7} 
                             & \begin{tabular}[c]{@{}c@{}}Obj\\ Det\end{tabular} & Depth                     & \begin{tabular}[c]{@{}c@{}}Obj\\ Det\end{tabular} & Depth                     & \begin{tabular}[c]{@{}c@{}}Obj\\ Det\end{tabular} & Depth                     \\ \hline
R                            & 0.265                                             & 0.108                     & 0.386                                             & 0.573                     & 0.349                                             & 0.319                     \\
G                            & 0.194                                             & 0.004                     & 0.236                                             & 0.251                     & 0.570                                             & 0.745                     \\
B                            & 0.137                                             & 0.001                     & 0.332                                             & 0.788                     & 0.531                                             & 0.211                     \\ \hline
\multicolumn{1}{l|}{Average} & \multicolumn{1}{r}{0.199}                         & \multicolumn{1}{r}{0.038} & \multicolumn{1}{r}{0.318}                         & \multicolumn{1}{r}{0.537} & \multicolumn{1}{r}{0.483}                         & \multicolumn{1}{r}{0.425} \\ \hline \hline
\end{tabular}%
}
\vspace{-5mm}
\end{table}

%% file: tables/obj_det_table_channel.tex
\begin{table}[]
\centering
\caption{Detection results for varying channel sizes}
\label{tab:channel-flir}
\resizebox{\columnwidth}{!}{%
\begin{tabular}{ccccccc}
\hline \hline
\multicolumn{7}{c}{Dataset: FLIR-ADAS}                                                                                                   \\ \hline \hline
\multicolumn{7}{c}{RetinaNet}                                                                                                            \\ \hline \hline
\multicolumn{1}{c|}{Compression}         & mAP           & \begin{tabular}[c]{@{}c@{}}mAP\\ @0.5\end{tabular}       & \begin{tabular}[c]{@{}c@{}}mAP\\ @0.75\end{tabular}      & \begin{tabular}[c]{@{}c@{}}mAP\\ @small\end{tabular}     & \begin{tabular}[c]{@{}c@{}}mAP\\ @medium\end{tabular}    & \begin{tabular}[c]{@{}c@{}}mAP\\ @large\end{tabular}     \\ \hline
\multicolumn{1}{c|}{$3 \rightarrow 3 $}  & 32.5          & 52.9          & 35.4          & 21.2          & 60.2          & 55.0          \\
\multicolumn{1}{c|}{$8 \rightarrow 3 $}  & 33            & 51.8          & 35.2          & 20.0          & 63.1          & \textbf{67.7} \\
\multicolumn{1}{c|}{$10 \rightarrow 3 $} & \textbf{34.7} & \textbf{53.3} & \textbf{37.4} & \textbf{22.7} & \textbf{64.1} & 56.5          \\
\multicolumn{1}{c|}{$12 \rightarrow 3 $} & 33.9          & 51.2          & 35.6          & 20.3          & 62.1          & 62.1          \\ \hline \hline
\multicolumn{7}{c}{YOLO-X}                                                                                                               \\ \hline \hline
\multicolumn{1}{c|}{Compression}         & mAP           & \begin{tabular}[c]{@{}c@{}}mAP\\ @0.5\end{tabular}       & \begin{tabular}[c]{@{}c@{}}mAP\\ @0.75\end{tabular}      & \begin{tabular}[c]{@{}c@{}}mAP\\ @small\end{tabular}     & \begin{tabular}[c]{@{}c@{}}mAP\\ @medium\end{tabular}    & \begin{tabular}[c]{@{}c@{}}mAP\\ @large\end{tabular}     \\ \hline
\multicolumn{1}{c|}{$3 \rightarrow 3 $}  & 54.8          & 83.3          & 58.2          & 47.0          & 72.3          & 72.1          \\
\multicolumn{1}{c|}{$8 \rightarrow 3 $}  & 55.5          & 83.0          & 59.8          & 47.4          & 73.2          & 74.9          \\
\multicolumn{1}{c|}{$10 \rightarrow 3 $} & 55.8          & 82.4          & 58.4          & 45.8          & 72.7          & 72.9          \\
\multicolumn{1}{c|}{$12 \rightarrow 3 $} & \textbf{56.0} & \textbf{83.8} & \textbf{60.5} & \textbf{47.8} & \textbf{73.4} & \textbf{77.0}          \\ \hline \hline
\multicolumn{7}{c}{Sparse RCNN}                                                                                                          \\ \hline \hline
\multicolumn{1}{c|}{Compression}         & mAP           & \begin{tabular}[c]{@{}c@{}}mAP\\ @0.5\end{tabular}       & \begin{tabular}[c]{@{}c@{}}mAP\\ @0.75\end{tabular}      & \begin{tabular}[c]{@{}c@{}}mAP\\ @small\end{tabular}     & \begin{tabular}[c]{@{}c@{}}mAP\\ @medium\end{tabular}    & \begin{tabular}[c]{@{}c@{}}mAP\\ @large\end{tabular}     \\ \hline
\multicolumn{1}{c|}{$3 \rightarrow 3 $}  & 45            & 69.8          & 49.7          & \textbf{36.4} & 65.3          & 57.6          \\
\multicolumn{1}{c|}{$8 \rightarrow 3 $}  & 45.6          & 69.7          & 49.0          & 35.1          & \textbf{69.2} & 65.7          \\
\multicolumn{1}{c|}{$10 \rightarrow 3 $} & 44.8          & 69.5          & 49.1          & 35.1          & 66.8          & \textbf{66.7} \\
\multicolumn{1}{c|}{$12 \rightarrow 3 $} & \textbf{45.7} & \textbf{70.5} & \textbf{49.8} & 36.1          & 67.5          & 62.7          \\ \hline \hline
\end{tabular}%
}
\end{table}

%% file: tables/indoor_depth_channel.tex
\begin{table}[h]
\centering
\caption{Depth estimation results for varying channel sizes}
\label{tab:depth_outdoor_channels}
\resizebox{\columnwidth}{!}{%
\begin{tabular}{c|cccc|ccc}
\hline \hline
\multirow{2}{*}{Compression}    & \multicolumn{4}{c|}{Error}                                        & \multicolumn{3}{c}{Accuracy}                                                                                                                                                        \\ \cline{2-8} 
                                & AbsRel         & SqRel          & RMS            & RMSlog         & \begin{tabular}[c]{@{}c@{}} $ \delta$\\ 1.25 \end{tabular} & \begin{tabular}[c]{@{}c@{}} $ \delta$\\ 1.25$^2$ \end{tabular} & \begin{tabular}[c]{@{}c@{}} $ \delta$\\ 1.25$^3$ \end{tabular} \\ \hline
$3 \rightarrow 3 $  & \textbf{0.098} & \textbf{0.611} & \textbf{3.683} & \textbf{0.136} & \textbf{0.915}                                            & \textbf{0.985}                                             & \textbf{0.995}                                             \\
$8 \rightarrow 3 $  & 0.109          & 0.725          & 4.038          & 0.148          & 0.976                                                     & 0.977                                                      & 0.993                                                       \\
$10 \rightarrow 3 $ & 0.109          & 0.682          & 4.029          & 0.15           & 0.884                                                     & 0.974                                                      & 0.992                                                      \\
$12 \rightarrow 3 $ & 0.109          & 0.747          & 4.084          & 0.148          & 0.891                                                     & 0.978                                                      & 0.993                                                      \\ \hline \hline
\end{tabular}%
}
\end{table}

%% file: tables/detection-sensitivity.tex
\begin{table}[]
\centering
\vspace{-3mm}
\caption{Sensitivity of the model to random D values. }
\resizebox{\columnwidth}{!}{%
\begin{tabular}{ccccc}
\hline \hline
\multicolumn{5}{c}{Dataset: FLIR-ADAS}                                                                                                                                                                                                                                                                                \\ \hline \hline
\multicolumn{5}{c}{Model: YOLO-X}                                                                                                                                                                                                                                                                                     \\ \hline \hline
\multicolumn{1}{c|}{Compression}             & \begin{tabular}[c]{@{}c@{}}max mAP \\ over 10 runs\end{tabular} & \begin{tabular}[c]{@{}c@{}}min mAP \\ over 10 runs\end{tabular} & \begin{tabular}[c]{@{}c@{}}mean mAP\\ over 10 runs\end{tabular} & \begin{tabular}[c]{@{}c@{}}s.t.d. mAP\\ over 10 runs\end{tabular} \\ \hline
\multicolumn{1}{c|}{TE(3)}  & 54.6                                                            & 54.3                                                            & 54.45                                                           & $8.50 \times 10^{-4}$                                             \\
\multicolumn{1}{c|}{TE(10)} & 52.2                                                            & 52.5                                                            & 52.36                                                           & $9.66 \times 10^{-4}$                                             \\
\multicolumn{1}{c|}{$3 \rightarrow 3$}      & 54.8                                                            & 54.5                                                            & 54.57                                                           & $9.49 \times 10^{-4}$                                             \\
\multicolumn{1}{c|}{$8 \rightarrow 3$}      & 55.5                                                            & 55.3                                                            & 55.42                                                           & $7.89 \times 10^{-4}$                                             \\
\multicolumn{1}{c|}{$10 \rightarrow 3$}     & 55.8                                                            & 53.9                                                            & 54.53                                                           & $7.8 \times 10^{-4}$                                              \\
\multicolumn{1}{c|}{$12 \rightarrow 3$}     & \textbf{56.0}                                                              & \textbf{55.6}                                                            & \textbf{55.85}                                                         & \textbf{$1.35\times 10^{-4}$}                                              \\ \hline \hline
\end{tabular}%
}
\label{tab:random_sensitivity}
\end{table}

%% file: tables/flops_table.tex
\begin{table}[]
\centering
\vspace{-5mm}
\caption{\bl{FLOPs evaluation on TCNet-SparseRCNN}}
\label{tab:flops}
\resizebox{\columnwidth}{!}{%
\begin{tabular}{cccc}
\hline
\textbf{Num channels} & \textbf{\begin{tabular}[c]{@{}c@{}}Thermal Embedding\\ Runtime\\ (ms)\end{tabular}} & \textbf{\begin{tabular}[c]{@{}c@{}}Sparse-RCNN\\ FLOPs\\ (FLOPs)\end{tabular}} & \textbf{\begin{tabular}[c]{@{}c@{}}Extra \\ FLOPs\\ (FLOPs)\end{tabular}} \\ \hline
None                  & 0                                                                                & 53.26G                                                                         & 0                                                                         \\
TE(3)                 & 4.35                                                                                & 53.26G                                                                         & 0                                                                         \\
TE(10)                & 25.37                                                                               & 55.06G                                                                         & 1.8G                                                                      \\ \hline
TC-3                  & 4.35                                                                                & 53.27G                                                                         & 29.12M                                                                    \\
TC-8                  & 19.71                                                                               & 53.3G                                                                          & 77.66M                                                                    \\
TC-10                 & 25.37                                                                               & 53.31G                                                                         & 97.08M                                                                    \\
TC-12                 & 31.41                                                                               & 53.32G                                                                         & 116.5M                                                                    \\ \hline
\end{tabular}%
}
\end{table}

%% file: tables/color-aug.tex
\begin{table}[h]
\caption{Detection results for different augmentations}
\label{color-augmentation}
\resizebox{\columnwidth}{!}{%
\begin{tabular}{ccccccc}
\hline \hline
Augmentation      & mAP            & \begin{tabular}[c]{@{}c@{}}mAP\\ @0.5\end{tabular} & \begin{tabular}[c]{@{}c@{}}mAP\\ @0.75\end{tabular} & \begin{tabular}[c]{@{}c@{}}mAP\\ @small\end{tabular} & \begin{tabular}[c]{@{}c@{}}mAP\\ @medium\end{tabular} & \begin{tabular}[c]{@{}c@{}}mAP\\ @large\end{tabular} \\ \hline
No Aug            & 0.379          & 0.63                                               & 0.397                                               & 0.311                                                & 0.528                                                 & 0.586                                                \\
HSV augmentation    & 0.362          & 0.583                                              & 0.401                                               & 0.271                                                & 0.552                                                 & 0.617                                                \\
Brightness jitter & 0.369          & 0.585                                              & 0.408                                               & 0.29                                                 & 0.553                                                 & 0.618                                                \\
Inversion         & 0.352          & 0.577                                              & 0.379                                               & 0.242                                                & 0.565                                                 & \textbf{0.647}                                       \\
Gamma correction  & 0.35           & 0.559                                              & 0.389                                               & 0.261                                                & 0.558                                                 & 0.641                                                \\
Thermal Embedding & 0.416          & 0.653                                              & 0.45                                                & 0.325                                                & 0.611                                                 & 0.637                                                \\
TCNet             & \textbf{0.457} & \textbf{0.705}                                     & \textbf{0.498}                                      & \textbf{0.361}                                       & \textbf{0.675}                                        & 0.627                                                \\ \hline \hline
\end{tabular}%
}
\end{table}

%% file: conclusion.tex
\section{Conclusion and Future works}
\label{sec:conclusion}


In this paper, we introduced Thermal Chameleon Network (TCNet), a task-adaptive tone-mapping framework for radiometric Thermal Infrared (TIR) images. TCNet's capability to simulate unseen TIR environments through thermal embedding and adaptive channel compression has demonstrated robust performance in object detection and depth estimation.

\bl{However, TCNet currently relies on radiometric TIR images, limiting its applicability to non-radiometric datasets. Additionally, unlike the baselines, TCNet requires task and model specific training, necessitating separate models for different tasks. Future work should focus on adapting the thermal embedding for non-radiometric RAW TIR images and developing a more generalized version for broader use.}